\documentclass[letterpaper]{article} 

\usepackage{aaai2026}  

\usepackage{times}  
\usepackage{helvet}  
\usepackage{courier}  
\usepackage[hyphens]{url}  
\usepackage{graphicx} 
\usepackage{natbib}  
\usepackage{caption} 
\usepackage{algorithm}
\usepackage{algorithmic}

\usepackage{newfloat}
\usepackage{listings}
\DeclareCaptionStyle{ruled}{labelfont=normalfont,labelsep=colon,strut=off} 
\floatstyle{ruled}
\newfloat{listing}{tb}{lst}{}
\floatname{listing}{Listing}
\usepackage[colorlinks=true]{hyperref} 

\usepackage{makecell,pifont,multirow}
\usepackage{colortbl}
\usepackage{caption}
\usepackage{booktabs}
\usepackage{subcaption}
\usepackage{amssymb}
\usepackage{marvosym}
\usepackage{titletoc}
\usepackage{amsmath}
\usepackage{xcolor}

\def\netName{VeteranAD}

\title{Perception in Plan: Coupled Perception and Planning for End-to-End Autonomous Driving}
\author {
    Bozhou Zhang\textsuperscript{\rm 1}\equalcontrib\quad
    Jingyu Li\textsuperscript{\rm 1,2}\equalcontrib\quad
    Nan Song\textsuperscript{\rm 1}\quad
    Li Zhang\textsuperscript{\rm 1,2}\corrauthor
}
\affiliations {
    \textsuperscript{\rm 1}School of Data Science, Fudan University\quad\quad
    \textsuperscript{\rm 2}Shanghai Innovation Institute\\
    \vspace{.4em} 
}

\begin{document}

\maketitle


\begin{abstract}

End-to-end autonomous driving has achieved remarkable advancements in recent years.
Existing methods primarily follow a perception–planning paradigm, where perception and planning are executed sequentially within a fully differentiable framework for planning-oriented optimization.
We further advance this paradigm through a ``perception-in-plan'' framework design, which integrates perception into the planning process. 
This design facilitates targeted perception guided by evolving planning objectives over time, ultimately enhancing planning performance.
Building on this insight, we introduce \textbf{\netName{}}, a coupled perception and planning framework for end-to-end autonomous driving.
By incorporating multi-mode anchored trajectories as planning priors, the perception module is specifically designed to gather traffic elements along these trajectories, enabling comprehensive and targeted perception.
Planning trajectories are then generated based on both the perception results and the planning priors.
To make perception fully serve planning, we adopt an autoregressive strategy that progressively predicts future trajectories while focusing on relevant regions for targeted perception at each step.
With this simple yet effective design, \netName{} fully unleashes the potential of planning-oriented end-to-end methods, leading to more accurate and reliable driving behavior.
Extensive experiments on the NAVSIM and Bench2Drive datasets demonstrate that our \netName{} achieves state-of-the-art performance.

\end{abstract}

\section{Introduction}
\label{sec:intro}


\begin{figure}[t!]
\centering

\includegraphics[width=0.47\textwidth]{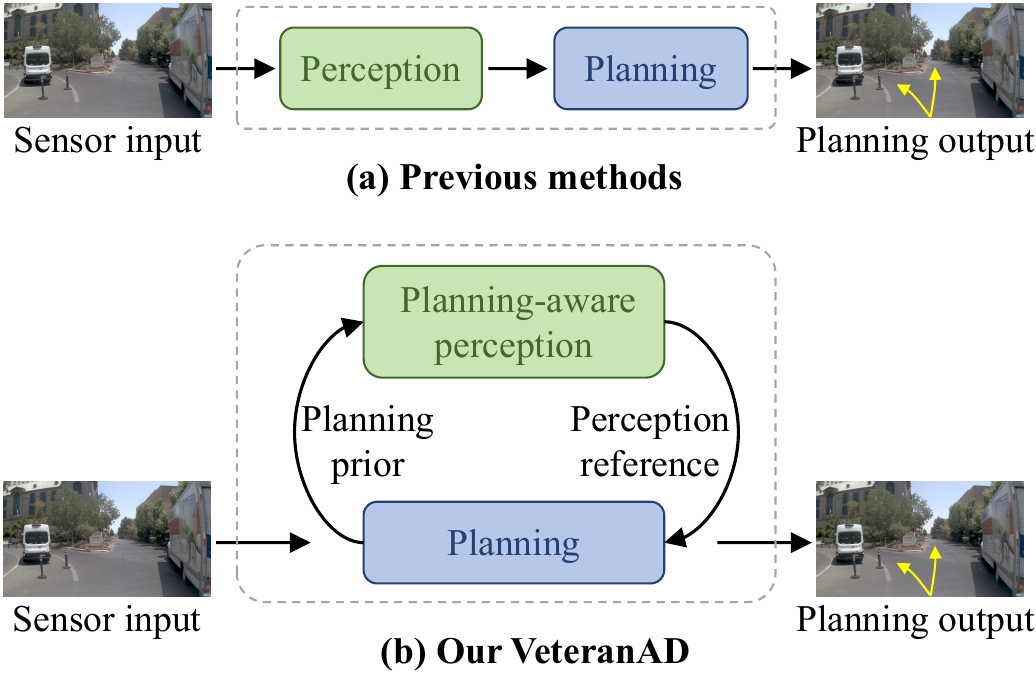}

\caption{
Comparison of end-to-end autonomous driving methods.
\textbf{(a)} Previous approaches mainly follow a perception–planning paradigm, executing these modules sequentially.
\textbf{(b)} In contrast, our \netName{} \textit{integrates perception into the planning process}, using planning priors to guide perception and leveraging targeted perception results to inform planning.
This ``perception-in-plan" paradigm enhances the planning-oriented framework.
For clarity, other intermediate modules are omitted in the figure.
}
\label{fig:first}
\end{figure}

End-to-end autonomous driving~\cite{uniad,vad,Transfuser} has made significant progress in recent years by unifying multiple tasks, including perception~\cite{bevformer,streampetr,MapTR}, prediction~\cite{mtr,qcnet}, and planning~\cite{tuplan,plantf}, into a unified framework.
In this way, the end-to-end driving framework builds a fully differentiable learning system that ensures planning-oriented optimization. This enables impressive performance in both open-loop~\cite{genad,sparsedrive,bridgead} and closed-loop~\cite{driveadapter,thinktwice,drivetransformer} planning.

Mainstream end-to-end autonomous driving methods~\cite{uniad,vad,sparsedrive,TransfuserPAMI} typically adopt a sequential paradigm, where perception is followed by planning, as shown in Figure~\ref{fig:first} (a). A Transformer-based architecture~\cite{attention} is often employed to make the entire pipeline differentiable, thereby enabling planning-oriented optimization.
However, differentiability alone is insufficient to fully exploit the advantages of planning-oriented optimization in end-to-end autonomous driving, whose goal is to ensure that all preceding modules—such as perception—are optimized to better serve the planning process.

To address the aforementioned limitation, we propose a ``perception-in-plan'' paradigm, which integrates perception into the planning process. In this way, the perception module operates in a targeted manner, aligned with the requirements of the planning process. Based on this paradigm, we introduce \textbf{\netName{}}, as illustrated in Figure~\ref{fig:first} (b).
In our framework, perception and planning are tightly coupled. Multi-mode anchored trajectories are used as planning priors to guide the perception module in gathering traffic elements—such as lanes and surrounding agents—along the predicted trajectories, enabling holistic and targeted perception for planning.
To fully inject perception into planning, we adopt an autoregressive strategy that progressively generates future trajectories. At each time step, guided by planning priors, the model focuses on relevant regions to perform targeted perception and generate the planning output for the corresponding step.
Under this paradigm, we design two core modules: Planning-Aware Holistic Perception and Localized Autoregressive Trajectory Planning.
The Planning-Aware Holistic Perception module operates across three dimensions: image features, bird’s-eye-view (BEV) features, and surrounding agent features. This interaction enables a comprehensive understanding of traffic elements, including vehicles, lanes, and barriers.
The Localized Autoregressive Trajectory Planning module decodes future trajectories in an autoregressive manner. It iteratively adjusts the anchor trajectories from near to far future based on perception results, ensuring context-aware and progressively refined planning.
Through the above design, \netName{} leverages trajectory priors to enable focused perception and progressive planning, thereby achieving strong end-to-end planning performance.

Our \textbf{contributions} are summarized as follows:
(i) We propose \netName{}, a novel framework that follows a ``perception-in-plan'' paradigm, integrating perception into the planning process.
(ii) We design two key modules: the Planning-Aware Holistic Perception module and the Localized Autoregressive Trajectory Planning module, which jointly couple perception and planning, fully unleashing the planning-oriented optimization advantages enabled by end-to-end autonomous driving.
(iii) Extensive experiments on the NAVSIM and Bench2Drive datasets demonstrate that \netName{} achieves state-of-the-art performance.

\section{Related work}


\begin{figure*}[t!]
\centering

\includegraphics[width=1\textwidth]{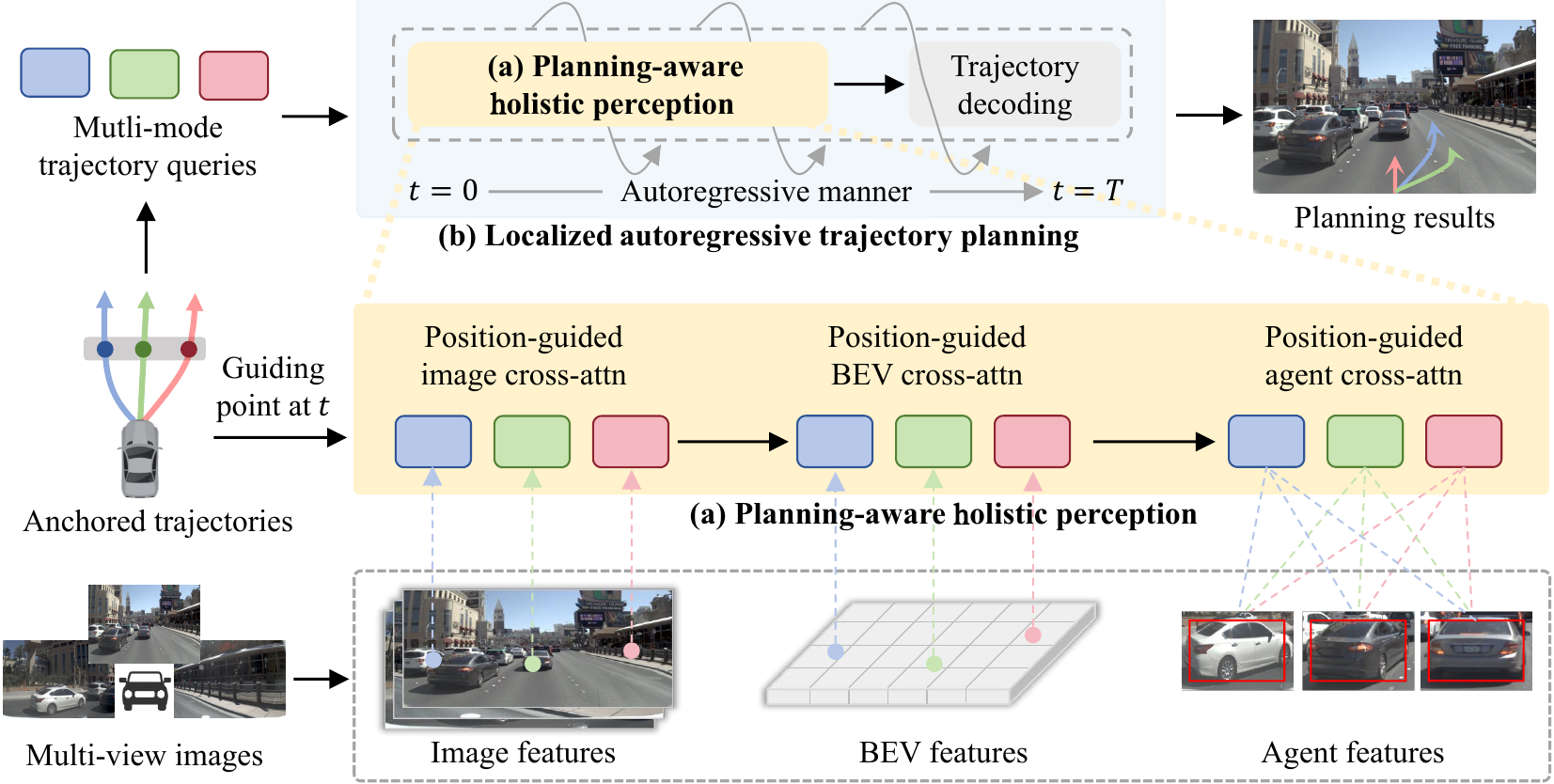}

\caption{
Overview of the \textbf{\netName{}} framework.
Multi-mode planning queries are initialized from the anchored trajectories.
And multi-view images are processed by the encoder to generate image features, BEV features, and agent features.
\textbf{(a)} The Planning-Aware Holistic Perception module takes the planning queries and performs cross-attention with image features, BEV features, and surrounding agent features, guided by the points on the anchored trajectories.
\textbf{(b)} 
The Localized Autoregressive Trajectory Planning module adjusts the planning trajectories, derived from the anchored trajectories, using the outputs from (a). It performs perception at each time step to generate the corresponding planning waypoint in an autoregressive manner, resulting in the complete planning trajectory.
}
\label{fig:main}
\end{figure*}

\paragraph{End-to-end autonomous driving.}

In the early stages of autonomous driving, rule-based methods adopted a modular design~\cite{congested,stanley,odin,parting}, dividing the system into separate components—perception, prediction, planning, and control—connected via predefined rules. While interpretable, this architecture suffers from error propagation and limited scenario coverage.
End-to-end planning methods~\cite{Transfuser,diffusiondrive} gradually replace individual modules, such as perception and planning, with deep learning-based sub-networks, while retaining essential rule-based constraints. This paradigm has gained traction for its ability to integrate perception, prediction, and planning into a unified framework, removing the need for hand-crafted intermediate representations.
Early works~\cite{Transfuser,TransfuserPAMI} typically bypassed intermediate tasks like perception and motion prediction. ST-P3~\cite{st-p3} was the first to introduce explicit intermediate representations in a surround-view camera-based framework. UniAD~\cite{uniad} further unified perception, prediction, and planning using transformer-based query interactions, achieving strong performance on the nuScenes~\cite{nuscenes} benchmark~\cite{bridgead,paradrive,song2025don,chen2024ppad,su2024difsd,doll2024dualad,ye2023fusionad,sun2025focalad,wang2025cogad}.
Recent advances explore diverse representations and learning paradigms. VAD~\cite{vad} proposed vectorized scene representations, while VADv2~\cite{vadv2} introduced probabilistic planning with a 4K trajectory vocabulary and conflict-aware loss, yielding state-of-the-art closed-loop performance on CARLA Town05~\cite{driveadapter,thinktwice}.
SparseDrive~\cite{sparsedrive} enhances efficiency through sparse scene representations and a parallel motion planner. GenAD~\cite{genad} adopts a generative framework that unifies motion prediction and planning using an instance-centric scene representation and structured latent modeling via variational autoencoders.
Recently, with the introduction of more challenging real-world benchmarks~\cite{navsim} and closed-loop simulation-based benchmarks~\cite{Bench2Drive,zimmerlin2024hidden} built on CARLA~\cite{carla}, an increasing number of works have explored various approaches for end-to-end autonomous driving, such as diffusion policy~\cite{diffusiondrive,jiang2025transdiffuser}, vision-language models~\cite{jiang2024senna,tian2024drivevlm,zeng2025futuresightdrive}, pure transformer architecture~\cite{drivetransformer}, reinforcement learning~\cite{rad,alphadrive,recogdrive,reinforced,li2025finetuning,chen2025rift}, closed-loop simulations,~\cite{yang2024drivearena} vision-language-action models~\cite{orion,chi2025impromptu,renz2025simlingo,zhou2025autovla}, dual systems~\cite{hamdan2025eta}, mixture of experts~\cite{yang2025drivemoe}, flow matching~\cite{xing2025goalflow}, test-time training~\cite{centaur}, bridging the gap between open-loop
training and closed-loop deployment~\cite{li2025hydra,tang2025hip}, trajectory selection~\cite{yao2025drivesuprim}, iterative planning~\cite{guo2025ipad}, and world models~\cite{wote,zheng2025world4drive,SSR,lienhancing,wang2024driving,chen2024drivinggpt}.
These existing methods primarily follow a ``perception–planning'' paradigm, aiming to improve performance by separately enhancing perception and planning capabilities. In contrast, our \netName{} adopts a ``perception-in-plan'' paradigm, which integrates perception directly into the planning process to enable more effective, planning-oriented optimization.

\paragraph{Closed-loop and open-loop benchmarking.}

Closed-loop and open-loop benchmarks are commonly used to evaluate autonomous driving systems.
Closed-loop evaluation simulates the full feedback loop—from sensor input to control execution—using tools such as nuPlan~\cite{nuPlan}, Waymax~\cite{waymax}, CARLA~\cite{carla}, Bench2Drive~\cite{Bench2Drive}, and MetaDrive~\cite{metadrive}.
These simulators enable measurement of driving metrics such as collision rate and ride comfort.
However, simulating realistic traffic behavior and sensor data remains challenging. Graphics-based rendering introduces domain gaps~\cite{neuroncap}, while data-driven sensor simulation suffers from limited visual quality~\cite{amini2020learning,vista,wang2022learning}.
Open-loop evaluation tests trajectory prediction on offline datasets like nuScenes~\cite{nuscenes}, without interaction with the environment.
Due to the lack of standardized planning metrics, prior works often rely on custom implementations, leading to inconsistent results~\cite{paradrive, li2024ego}.

\section{Methodology}

\subsection{Preliminary}

\paragraph{Task formulation.}
End-to-end autonomous driving takes sensor data (such as camera and LiDAR) as input and generates future planning trajectories as output. The planning task typically involves generating multi-mode trajectories to represent multiple possible future plans. Auxiliary tasks, such as detection, map segmentation, and motion prediction for surrounding agents, are also integrated into the end-to-end models to help the model better learn scene features for safe planning results.

\paragraph{Framework overview.}
The framework of our \netName{} is illustrated in Figure~\ref{fig:main}. It comprises three main components: an image encoder, the Planning-Aware Holistic Perception module, and the Localized Autoregressive Trajectory Planning module.
First, the image encoder extracts features from multi-view images, producing image features, BEV features, and surrounding agent features.
Next, multi-mode trajectory queries are initialized from the anchored trajectories.
The Planning-Aware Holistic Perception module performs position-guided cross-attention between the trajectory queries and the extracted image, BEV, and agent features.
The Localized Autoregressive Trajectory Planning module then operates in an autoregressive manner, performing perception at each time step and adjusting the anchored trajectory point accordingly, ultimately generating the complete planning output.

\subsection{Image encoding}
Given multi-view images $I \in \mathbb{R}^{N_{\rm img} \times 3 \times H \times W}$, where $N_{\rm img}$ denotes the number of camera views, the image encoder~\cite{resnet} first extracts multi-view image features, denoted as $F_{\rm img}$.
Then, bird’s-eye-view (BEV) features $F_{\rm BEV}$ are generated from image features using the LSS~\cite{lss} method. A simple multi-layer perceptron (MLP) decoder is then applied to decode the BEV features into a BEV segmentation map, which is supervised using the ground truth segmentation map.
The surrounding agent features $F_{\rm agent}$ are initialized and interact with BEV features through Transformer~\cite{attention} blocks. A simple MLP decoder then decodes the agent features into bounding boxes, which are supervised using the ground truth bounding boxes of the surrounding agents. The process is shown below:

\begin{equation}
\begin{split}
F_{\rm img} &= {\rm ImageEncoder}(I), \\
F_{\rm BEV} &= {\rm ImageToBEV}(F_{\rm img}), \\
F_{\rm agent} &= {\rm Transformer}({\rm Q} = F_{\rm agent}, {\rm K, V} = F_{\rm BEV}).
\end{split}
\end{equation}

After obtaining these features, the multi-mode trajectory queries $Q_{\rm traj} \in \mathbb{R}^{M \times C}$ are initialized from the anchored trajectories, where $M$ denotes the number of planning modes and $C$ represents the feature channels. The anchored trajectories are clustered from the ground truth planning trajectories following previous work~\cite{sparsedrive,diffusiondrive}.

\subsection{Planning-aware holistic perception}
The perception module enables the trajectory queries to comprehensively capture the scene and traffic elements, such as lanes, vehicles, pedestrians, and barriers, ensuring accurate and safe planning. Given the trajectory queries $Q_{\rm traj} \in \mathbb{R}^{M \times C}$, three types of cross-attention are employed to interact with image features, BEV features, and agent features.

The position-guided image cross-attention and position-guided BEV cross-attention are designed to selectively gather features along potential planning trajectories. First, the guiding points $P_{\rm t} \in \mathbb{R}^{M \times 3}$ at time $t$ are extracted from the anchored trajectories, which serve as the planning prior. These guiding points are then projected onto both the image and BEV coordinates. Following previous works~\cite{detr3d,sparsebev}, they serve as reference points for cross-attention between the trajectory queries and the image and BEV features. The process is shown below:

\begin{equation}
\begin{split}
Q_{\rm traj} &= {\rm CrossAttn}({\rm Q} = Q_{\rm traj}, {\rm K, V} = F_{\rm img}), \\
Q_{\rm traj} &= {\rm CrossAttn}({\rm Q} = Q_{\rm traj}, {\rm K, V} = F_{\rm BEV}).
\end{split}
\end{equation}

The position-guided agent cross-attention is designed to effectively differentiate the importance of surrounding agents based on their distance. As introduced in the image encoding section, bounding boxes are decoded, allowing the positions of agents to be obtained. 
The pairwise relative distances between the surrounding agents and the ego agent are then computed using the guiding points and the decoded agent positions.

The relative distances are first encoded by an MLP to obtain the relative distance feature $F_{\rm DisRel}$. This feature is then concatenated with the agent features and trajectory queries to form the distance-aware agent features $F_{\rm AgRel}$. Inspired by previous works~\cite{qcnet,simpl}, cross-attention is subsequently applied to enable interaction between the trajectory queries and the distance-aware agent features, after aligning their dimensions. The overall process is illustrated below:
\begin{equation}
\begin{split}
F_{\rm AgRel} &= {\rm Concat}(Q_{\rm traj}, F_{\rm agent}, F_{\rm DisRel}), \\
Q_{\rm traj} &= {\rm CrossAttn}({\rm Q} = Q_{\rm traj}, {\rm K, V} = F_{\rm AgRel}).
\end{split}
\end{equation}

\subsection{Localized autoregressive trajectory planning}
The trajectory planning module aims to use the anchored trajectories as coarse planning trajectories and incorporate scene features to generate the final planning trajectories. 
For the anchored multi-mode trajectories over the future $T$ steps, we obtain the trajectory points set $\{P_{\rm 1}, ..., P_{\rm T}\}$, where $P_{\rm t} \in \mathbb{R}^{M \times 3}$ is the same as mentioned above at time $t$. These trajectory points serve as guiding points for trajectory planning.
The process operates in an autoregressive manner. At each time step $t$, the module takes the trajectory queries $Q_{\rm traj}$ and the guiding point  $P_{\rm t}$ as input, while the Planning-Aware Holistic Perception module interacts with the trajectory queries and the scene features.
Then, an MLP trajectory decoder is employed to predict the future trajectory point at time step $t$. The model estimates the offset $\Delta P_{\rm ft}$ to refine the guiding point, producing the final planned trajectory point $P_{\rm ft}$, as shown below:

\begin{equation}
P_{\rm ft} = \Delta P_{\rm ft} + P_{\rm t}.
\end{equation}

Finally, we obtain the planned trajectory point set $\{P_{\rm f1}, ..., P_{\rm fT}\}$, which forms the final planning trajectories $P_{\rm f} \in \mathbb{R}^{M \times T \times 3}$. At the final time step $T$, the module decodes the classification score $S_{\rm f} \in \mathbb{R}^{M \times 1}$ for the multi-mode trajectories. 
To model the movement of trajectory points, we employ Motion-Aware Layer Normalization~\cite{streampetr} to transform trajectory queries from time $t-1$ to time $t$, conditioned on the guiding points at time $t$, inspired by previous works~\cite{streampetr,RealMotion}.

\subsection{End-to-end learning}
The loss function consists of four components: the BEV segmentation map loss $\mathcal{L}_{\rm BEV}$, the agent bounding box loss $\mathcal{L}_{\rm agent}$, the planning regression loss $\mathcal{L}_{\rm reg}$, and the planning classification loss $\mathcal{L}_{\rm cls}$. The BEV segmentation map loss is calculated using cross-entropy loss. The agent bounding box loss is divided into L1 loss for box position regression and binary cross-entropy loss for box label classification. The planning regression loss is L1 loss, while the planning classification loss is computed using Focal loss. The overall loss function for end-to-end training is as follows:

\begin{equation}
\mathcal{L}_{total}=\lambda_{1}\mathcal{L}_{BEV}+\lambda_{2}\mathcal{L}_{agent}+\lambda_{3}\mathcal{L}_{reg}+\lambda_{4}\mathcal{L}_{cls},
\end{equation}
where $\lambda_{1}$, $\lambda_{2}$, $\lambda_{3}$, and $\lambda_{4}$ are the balancing factors.

\section{Experiments}


\begin{table*} [ht!]
\centering
{\begin{tabular}{l|c|cccccc}
\toprule[1.5pt]
Method & Input & NC $\uparrow$ &DAC $\uparrow$ & TTC $\uparrow$& Comf. $\uparrow$ & EP $\uparrow$ & \cellcolor{gray!30}PDMS $\uparrow$  \\
\midrule
VADv2-$\mathcal{V}_{8192}$~\cite{vadv2} & C \& L & 97.2 & 89.1 & 91.6 & \textbf{100} & 76.0 & \cellcolor{gray!30}80.9 \\
Hydra-MDP-$\mathcal{V}_{8192}$~\cite{hydramdp} & C \& L & 97.9 & 91.7 & 92.9 & \textbf{100} & 77.6 & \cellcolor{gray!30}83.0 \\
UniAD~\cite{uniad} & Camera & 97.8 & 91.9 & 92.9 & \textbf{100} & 78.8 & \cellcolor{gray!30}83.4 \\
LTF~\cite{Transfuser} & Camera & 97.4 & 92.8 & 92.4 & \textbf{100} & 79.0 & \cellcolor{gray!30}83.8 \\
PARA-Drive~\cite{paradrive} & Camera & 97.9 & 92.4 & 93.0 & 99.8 & 79.3 & \cellcolor{gray!30}84.0 \\
Transfuser~\cite{Transfuser} & C \& L & 97.7 & 92.8 & 92.8 & \textbf{100} & 79.2 & \cellcolor{gray!30}84.0 \\
DRAMA~\cite{drama} & C \& L & 98.0 & 93.1 & 94.8 & \textbf{100} & 80.1 & \cellcolor{gray!30}85.5 \\

Hydra-MDP++~\cite{HydraMDPplus} & Camera & 97.6 & 96.0 & 93.1 & \textbf{100} & 80.4 & \cellcolor{gray!30}86.6 \\

DiffusionDrive~\cite{diffusiondrive} & C \& L & 98.2 & 96.2 & 94.7 & \textbf{100}  & \underline{82.2}  & \cellcolor{gray!30}88.1\\
WoTE~\cite{wote}                     & C \& L & \underline{98.5} & \underline{96.8} & \underline{94.9} & \underline{99.9} & 81.9 & \cellcolor{gray!30}\underline{88.3} \\

\rowcolor{gray!30}
\bf\netName~(Ours) & Camera & \bf99.1 & \bf98.3 & \bf96.1 & \bf100 & \bf83.1 & \cellcolor{gray!30}\bf90.2 \\

\bottomrule[1.5pt]
\end{tabular}}
\caption{
Performance comparison \textit{on the NAVSIM~\cite{navsim} dataset for the navtest split} using closed-loop metrics.
``C \& L" indicates the use of both camera and LiDAR.
The \textbf{best} and \underline{second best} results are highlighted in \textbf{bold} and \underline{underline}.
}
\label{tab:navsim}
\end{table*}


\begin{table*} [ht!]
\centering

{
\begin{tabular}{l|c|cc}
\toprule[1.5pt]

\multirow{2}{*}{Method} & \multicolumn{1}{c|}{Open-loop} & \multicolumn{2}{c}{Closed-loop} \\ \cmidrule{2-4}
& \multicolumn{1}{c|}{Average L2 Error $\downarrow$} & \multicolumn{1}{c}{Driving Score $\uparrow$}  & \multicolumn{1}{c}{Success Rate (\%) $\uparrow$} \\ \midrule

AD-MLP~\cite{admlp} & \multicolumn{1}{c|}{3.64} & \multicolumn{1}{c}{18.05} & \multicolumn{1}{c}{0.00} \\ 

VAD~\cite{vad} & \multicolumn{1}{c|}{0.91} & \multicolumn{1}{c}{42.35} & \multicolumn{1}{c}{15.00} \\

Dual-AEB~\cite{dualaeb} & \multicolumn{1}{c|}{-} & \multicolumn{1}{c}{45.23} & \multicolumn{1}{c}{10.00} \\

UniAD-Tiny~\cite{uniad} & \multicolumn{1}{c|}{0.80} & \multicolumn{1}{c}{40.73} & \multicolumn{1}{c}{13.18} \\

UniAD-Base~\cite{uniad} & \multicolumn{1}{c|}{{0.73}} & \multicolumn{1}{c}{{45.81}} & \multicolumn{1}{c}{{16.36}} \\  
DriveTransformer~\cite{drivetransformer} & \multicolumn{1}{c|}{\underline{0.62}} & \multicolumn{1}{c}{\underline{63.46}} & \multicolumn{1}{c}{\textbf{35.01}} \\

\rowcolor{gray!30}
\bf\netName~(Ours) & \multicolumn{1}{c|}{\textbf{0.60}} & \multicolumn{1}{c}{\textbf{64.22}} & \multicolumn{1}{c}{\underline{33.85}} \\ \midrule

TCP*~\cite{TCP} & \multicolumn{1}{c|}{1.70} & \multicolumn{1}{c}{40.70} & \multicolumn{1}{c}{15.00} \\
TCP-ctrl*~\cite{TCP} & \multicolumn{1}{c|}{-} & \multicolumn{1}{c}{30.47} & \multicolumn{1}{c}{7.27} \\
TCP-traj*~\cite{TCP} & \multicolumn{1}{c|}{1.70} & \multicolumn{1}{c}{59.90} & \multicolumn{1}{c}{30.00} \\
ThinkTwice*~\cite{thinktwice} & \multicolumn{1}{c|}{0.95} & \multicolumn{1}{c}{62.44} & \multicolumn{1}{c}{31.23} \\
DriveAdapter*~\cite{driveadapter} & \multicolumn{1}{c|}{1.01} & \multicolumn{1}{c}{64.12} & \multicolumn{1}{c}{33.08} \\

\bottomrule[1.5pt]

\end{tabular}}

\caption{
Performance comparison \textit{on CARLA v2 using the Bench2Drive~\cite{Bench2Drive} benchmark} under both open-loop and closed-loop evaluations.
``*" denotes expert feature distillation.
} 
\label{tab:bench2drive}

\end{table*}

\subsection{Experimental settings}

\paragraph{Datasets.}
NAVSIM~\cite{navsim} is a large-scale real-world autonomous driving dataset designed for non-reactive simulation and benchmarking.
It focuses on challenging scenarios involving dynamic intention changes, while filtering out trivial cases such as stationary or constant-speed driving.
NAVSIM provides sensor data from cameras and LiDAR, along with annotated HD maps and object bounding boxes at 2 Hz.
It is split into two subsets: navtrain (1,192 scenarios) for training and validation, and navtest (136 scenarios) for testing.

Bench2Drive~\cite{Bench2Drive} is the first closed-loop evaluation benchmark tailored for end-to-end autonomous driving.
It addresses the limitations of traditional open-loop evaluations by offering a more realistic and interactive testing setup.
The training set contains 10,000 short clips generated in the CARLA v2~\cite{carla} simulator, while the evaluation set includes 220 independent short routes.

\paragraph{Evaluation metrics.}
For the NAVSIM dataset, we evaluate our method using the PDM Score (PDMS) as defined in the official benchmark. PDMS consists of several sub-scores: No At-Fault Collisions (NC), Drivable Area Compliance (DAC), Time-to-Collision (TTC), Comfort (Comf.), and Ego Progress (EP).
For the Bench2Drive dataset, we follow the official evaluation protocols. In open-loop evaluation, we use the Average L2 Error.
For closed-loop evaluation, we adopt the Driving Score and Success Rate. 
Further details are provided in the appendix.

\paragraph{Implementation details.}
The model is trained using 8 NVIDIA GeForce RTX 3090 GPUs, with a total batch size of 32 for 16 epochs.
The learning rate and weight decay are set to $2 \times 10^{-4}$ and $1 \times 10^{-4}$, respectively, and the model is optimized with AdamW.
For fair comparison, the image backbone follows prior works and adopts ResNet-34~\cite{resnet}.
The input consists of three images, front-right, front, and front-left, which are resized to $768 \times 432$.
The number of planning modes is set to 20.
The balancing factors for loss calculation, $\lambda_{1}$, $\lambda_{2}$, $\lambda_{3}$, and $\lambda_{4}$, are all set to 10 for NAVSIM and 1 for Bench2Drive.
The anchored trajectories are clustered using K-Means, following the same procedure as in previous works~\cite{sparsedrive,diffusiondrive}.


\begin{table*} [ht!]
\centering
{\begin{tabular}{cc|cccccc}
\toprule[1.5pt]
Holistic perception & Trajectory planning & NC $\uparrow$ &DAC $\uparrow$ & TTC $\uparrow$& Comf. $\uparrow$ & EP $\uparrow$ & PDMS $\uparrow$  \\
\midrule
& \checkmark & 98.5 & 96.0 & 94.9 & 100 & 80.8 & 87.5 \\
\checkmark &  & 98.6 & 96.7 & 95.2 & 100 & 81.9 & 88.4 \\

\rowcolor{gray!30}
\checkmark & \checkmark & 99.1 & 98.3 & 96.1 & 100 & 83.1 & 90.2 \\

\bottomrule[1.5pt]
\end{tabular}}
\caption{Ablation study on the components, including the Planning-Aware Holistic Perception module and the Localized Autoregressive Trajectory Planning module.}
\label{tab:abl_main}
\end{table*}


\begin{table} [ht!]
\centering
{\begin{tabular}{ccc|c}
\toprule[1.5pt]
Img-attn & BEV-attn & Agent-attn & PDMS $\uparrow$  \\
\midrule
& \checkmark & \checkmark & 89.7 \\
\checkmark &  & \checkmark & 89.1 \\
\checkmark & \checkmark &  & 89.4 \\

\rowcolor{gray!30}
\checkmark & \checkmark & \checkmark & 90.2 \\

\bottomrule[1.5pt]
\end{tabular}}
\caption{Ablation study on cross-attention with image, BEV, and agent features.}
\label{tab:abl_attn}
\end{table}


\begin{table} [t!]
\centering
{\begin{tabular}{c|cccc}
\toprule[1.5pt]
Decoding type &DAC $\uparrow$ & TTC $\uparrow$ & EP $\uparrow$ & PDMS $\uparrow$  \\
\midrule
NAR & 97.1 & 94.9 & 82.1 & 88.6 \\

\rowcolor{gray!30}
AR & 98.3 & 96.1 & 83.1 & 90.2 \\

\bottomrule[1.5pt]
\end{tabular}}
\caption{Ablation study on the planning process in autoregressive (AR) and non-autoregressive (NAR) modes within the Localized Autoregressive Trajectory Planning module.}
\label{tab:abl_arnar}
\end{table}

\subsection{Comparison with state of the art}
As shown in Table~\ref{tab:navsim}, our \netName{} is compared with state-of-the-art methods on the NAVSIM navtest split.
Using the same ResNet-34 backbone, \netName{} achieves a PDM Score (PDMS) of 90.2, significantly outperforming previous learning-based methods.
With only camera input, \netName{} surpasses UniAD by 6.8 PDMS, demonstrating its superior performance.
Even compared to top-performing methods such as DiffusionDrive and WoTE, \netName{} achieves higher scores across all evaluation metrics.
These results highlight the effectiveness of our ``perception-in-plan'' design for end-to-end planning.
We further evaluate our method on CARLA v2 using the Bench2Drive benchmark under both open-loop and closed-loop settings.
As shown in Table~\ref{tab:bench2drive}, \netName{} achieves an average L2 distance of 0.60 in the open-loop evaluation, outperforming all baselines.
In the closed-loop evaluation, \netName{} delivers competitive performance, on par with state-of-the-art methods such as DriveTransformer and DriveAdapter.
These strong results demonstrate the effectiveness and generalization capability of our approach.

\subsection{Ablation study}

\paragraph{Effects of components.}
Table~\ref{tab:abl_main} presents an ablation study of the Planning-Aware Holistic Perception module and the Localized Autoregressive Trajectory Planning module.
The results include the performance of the full model as well as the effect of each module individually.
In the first row, replacing position-guided attention with vanilla attention in the perception module leads to a drop in PDMS, highlighting the importance of using guiding points from anchored trajectories as planning priors.
The second row shows that removing the guiding points and directly outputting planning trajectories—rather than predicting offsets—also results in significant performance degradation.
These results suggest that the guiding points from anchored trajectories play a crucial role in accurate planning.
When both modules are applied simultaneously, as shown in the third row, the PDMS reaches 90.2, demonstrating their complementary strengths and overall effectiveness.

\paragraph{Effects of different attention types.}
We investigate the impact of different attention types applied to image features, BEV features, and agent features, with the results summarized in Table~\ref{tab:abl_attn}.
Removing any single attention mechanism leads to a performance drop, with the most significant decline observed when BEV attention is removed, highlighting the critical role of road information in planning.
Each type of attention captures interactions with specific traffic elements, such as lanes, surrounding agents, and static obstacles.
The combination of all three attention mechanisms yields the best performance, as shown in the last row.


\begin{table*} [ht!] 
\centering

\centering
{\begin{tabular}[b]{c|cccc|cccc|c}
\toprule[1.5pt]

\multirow{2}{*}{Method} & 
\multicolumn{4}{c|}{\textbf{L2 ($m$)} $\downarrow$} & 
\multicolumn{4}{c|}{\textbf{Col. Rate (\%)} $\downarrow$} & \multirow{2}{*}{FPS} \\
& 1$s$ & 2$s$ & 3$s$ & Avg. & 1$s$ & 2$s$ & 3$s$ & Avg. & \\
\midrule 

VAD~\cite{vad} & 0.41 & 0.70 & 1.05 & 0.72 & 0.07 & 0.17 & 0.41 & 0.22 & 4.5 \\  
\rowcolor{gray!30}
VAD~+~\netName~& 0.33$_{\textit{-0.08}}$ & 0.59$_{\textit{-0.11}}$ & 0.94$_{\textit{-0.11}}$ & 0.62$_{\textit{-0.10}}$ & 0.03$_{\textit{-0.04}}$ & 0.12$_{\textit{-0.05}}$ & 0.34$_{\textit{-0.07}}$ & 0.16$_{\textit{-0.06}}$ & 4.3 \\

\bottomrule[1.5pt]
\end{tabular}}

\caption{Performance comparison of open-loop planning results \textit{on the nuScenes~\cite{nuscenes} validation dataset}.}
\label{tab:abl_nuscenes}
\end{table*}


\begin{figure}[t!]
\centering

\includegraphics[width=0.47\textwidth]{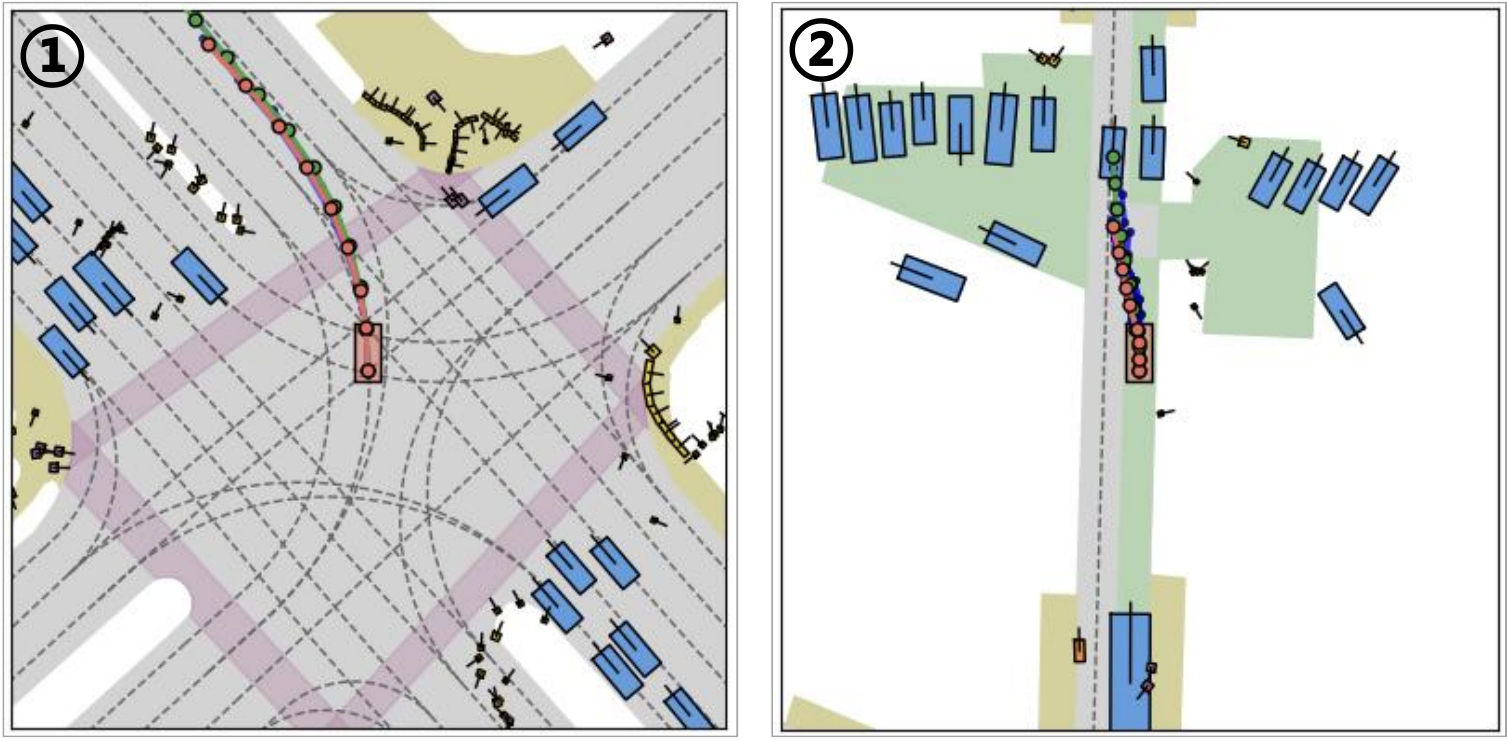}

\caption{
Qualitative results \textit{on the NAVSIM dataset}. The ground-truth trajectory is shown in green, and the planned trajectory is shown in orange.
}
\label{fig:main_navsim}
\end{figure}


\begin{figure*}[ht!]
\centering

\includegraphics[width=1\textwidth]{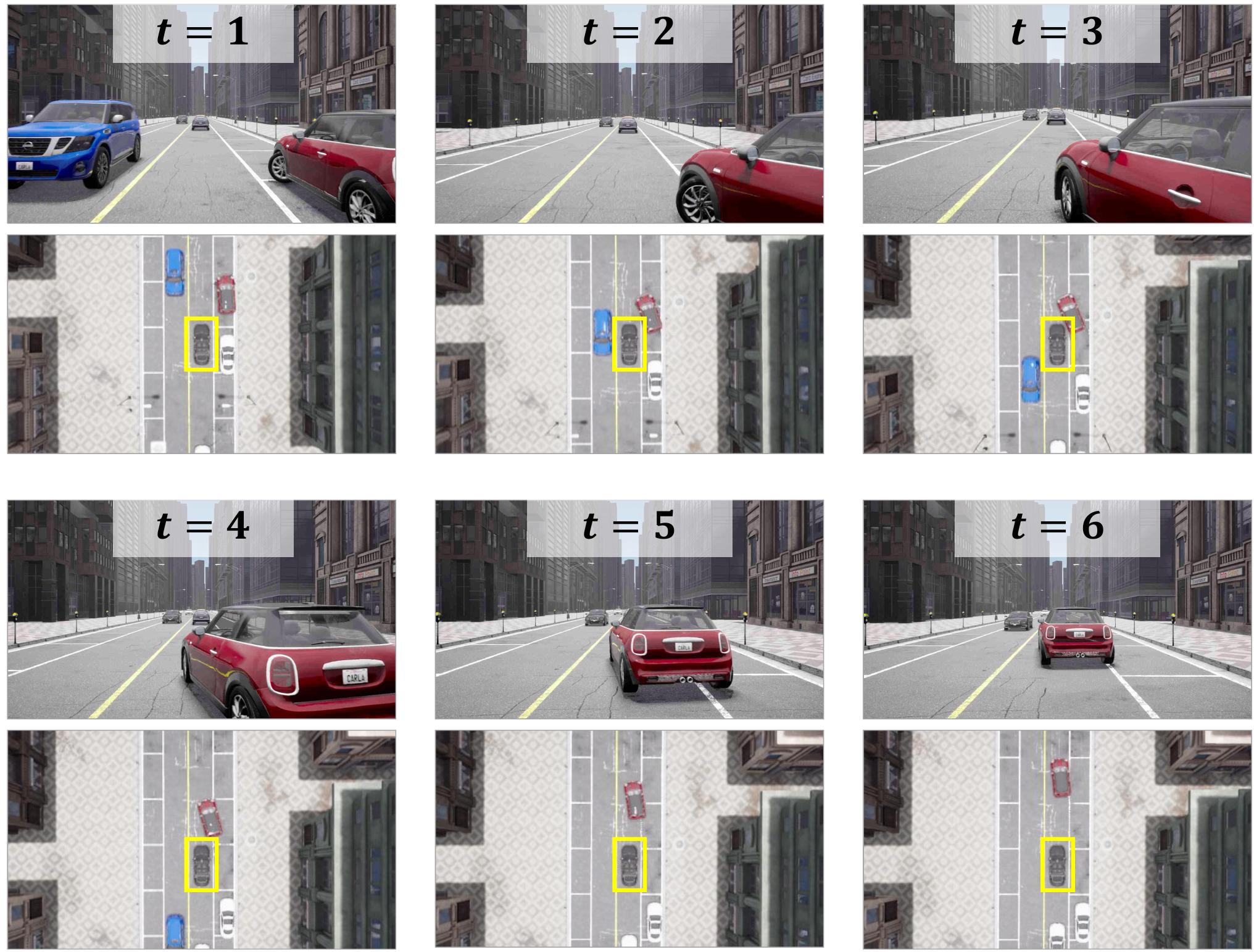}

\caption{
Qualitative results \textit{on the Bench2Drive dataset}.
The top image shows the front-view camera perspective, while the bottom image presents the BEV. In the BEV view, the ego vehicle is marked with a yellow box. In this scenario, our model successfully slows down and avoids a collision as a previously parked vehicle begins to merge into the road.
}
\label{fig:supp_ben}
\end{figure*}

\paragraph{AR $v.s.$ NAR.}
The autoregressive (AR) decoding process underpins the ``perception-in-plan'' framework by progressively predicting future trajectories while performing targeted perception at each step.
We conduct an ablation study by replacing the AR approach with a non-autoregressive (NAR) strategy in the trajectory planning module, as shown in Table~\ref{tab:abl_arnar}.
In the NAR setting, trajectory queries interact with scene features simultaneously in a one-shot manner, guided by all points in the anchored trajectories—thus following the conventional ``perception-planning'' paradigm.
The results show that the AR approach consistently outperforms the NAR counterpart. This is because, in the AR setting, trajectory queries focus on one trajectory point at a time, enabling step-wise adjustment and tighter coupling between perception and planning.
In contrast, the NAR method processes all points in parallel, making perception less responsive to planning intent and resulting in suboptimal performance.
This substantial improvement highlights the effectiveness of the AR-based, planning-oriented design in our framework.

\subsection{Comparison on nuScenes dataset}
To further validate the effectiveness and generalization of our \netName{}, we conduct an open-loop planning experiment on the nuScenes~\cite{nuscenes} dataset, with the results presented in Table~\ref{tab:abl_nuscenes}.
We integrate our design on top of VAD~\cite{vad} and follow its training and inference procedures. 
As shown, our approach reduces the average L2 Displacement Error by 0.10 m and lowers the average Collision Rate by 27.2\% compared to VAD.

\subsection{Efficiency analysis}
We compare our \netName{} with the state-of-the-art method DiffusionDrive~\cite{diffusiondrive}, following its official training and inference protocols.
Our model requires approximately 8 hours to train, compared to 9 hours for DiffusionDrive.
At inference time, \netName{} achieves an average latency of 22.3 ms, while DiffusionDrive runs at 18.4 ms.
Despite comparable efficiency in both training and inference, \netName{} achieves significantly better performance.
All experiments are conducted on NVIDIA GeForce RTX 3090 GPUs for fair comparison.

\subsection{Qualitative results}
As shown in Figure~\ref{fig:main_navsim}, on the NAVSIM dataset, our model accurately plans complex maneuvers such as left turns and lane changes.
As shown in Figure~\ref{fig:supp_ben}, during the closed-loop simulation on the Bench2Drive dataset, our model slows down to yield to a parked vehicle that begins merging into the road.
Additional visualizations and failure cases are provided in the appendix.

\section{Conclusion}
In this work, we propose \netName{}, an end-to-end autonomous driving framework built upon the proposed ``perception-in-plan'' paradigm, which tightly integrates perception into the planning process. By leveraging multi-mode anchored trajectories as planning priors, the Planning-Aware Holistic Perception module enables targeted and comprehensive understanding of the traffic scene, while the Localized Autoregressive Trajectory Planning module progressively refines future trajectories based on perception feedback. Extensive experiments on NAVSIM and Bench2Drive demonstrate that \netName{} achieves state-of-the-art performance.

\paragraph{Limitations and future work.}
The limitations of our model stem from its restricted capacity for closed-loop simulation, a common challenge for imitation learning-based end-to-end approaches. In the future, incorporating reinforcement learning may help enhance planning performance.

\newpage

\section*{Appendix}

\section{Discussions}
\paragraph{Discussion 1:}
\textbf{Key contributions of our work.}

The key contribution of our work is the introduction of a new paradigm for end-to-end autonomous driving, termed ``perception-in-plan'', which integrates the perception process into the planning process to fully leverage planning-oriented optimization. To realize this paradigm, we adopt an autoregressive strategy that performs planning progressively while conducting perception at each step. To ensure that perception is targeted and relevant at each step, we utilize anchored trajectories as planning priors to focus on specific regions. Based on this pipeline, we design two core modules—Planning-Aware Holistic Perception and Localized Autoregressive Trajectory Planning—that together constitute our \netName{} framework.

\paragraph{Discussion 2:}
\textbf{Relationship between Planning-Aware Holistic Perception and Localized Autoregressive Trajectory Planning.}

Localized Autoregressive Trajectory Planning performs planning in an autoregressive manner, while at each time step, Planning-Aware Holistic Perception conducts targeted perception.

\paragraph{Discussion 3:}
\textbf{Limitations and future work.}

In addition to the limitations and future work discussed in the conclusion section of the main paper, we provide a more in-depth discussion here. As shown in Table 2 of the main paper, although our \netName{} outperforms previous methods on the Bench2Drive dataset, it still achieves a relatively low success rate in closed-loop evaluation. This reflects a common challenge faced by imitation learning-based end-to-end approaches. In the future, incorporating reinforcement learning into the current end-to-end framework may significantly enhance the model’s ability to handle challenging scenarios.

\paragraph{Discussion 4:}
\textbf{About ``perception'' and ``planning''.}

Some works define the perception process as the extraction of image features, BEV features, and agent features from raw sensor data. In this paper, we extend the definition of perception to also include the interaction between trajectory queries and these extracted features.

Our key insight is that, with continued advancements in perception models, the extraction of such features has become increasingly reliable and accessible. As a result, the core challenge in end-to-end autonomous driving is no longer the extraction itself, but rather how to effectively model the relationship between these extracted scene features and the planning process—that is, the interface between perception and planning. Our \netName{} is specifically designed to address this challenge.

\section{Evaluation metrics}

\paragraph{NAVSIM.}
NAVSIM~\cite{navsim} scores driving agents in two steps. First, subscores in range $[0,1]$ are computed after simulation. Second, these subscores are aggregated into the PDM Score (PDMS) $\in[0,1]$. 
We use the following aggregation of subscores based on the official definition:

\begin{equation}
\begin{aligned}
\textrm{PDMS} = \underbrace{\Bigg( { \prod_{m \in \{\texttt{NC}, \texttt{DAC}\}}} \texttt{score}_m \Bigg)}_{\text{penalties}} \times \\
\underbrace{\Bigg( \frac{\sum_{w \in \{\texttt{EP}, \texttt{TTC}, \texttt{C}\}} \texttt{weight}_w \times  \texttt{score}_w}{\sum_{w \in \{\texttt{EP}, \texttt{TTC}, \texttt{C}\}} \texttt{weight}_w }  \Bigg)}_{\text{weighted average}}.
\end{aligned}
\end{equation}

Subscores are categorized by their importance as penalties or terms in a weighted average. A penalty punishes inadmissible behavior such as collisions with a factor $<1$. The weighted average aggregates subscores for other objectives such as progress and comfort.

\paragraph{Bench2Drive.}
We evaluate the performance using two closed-loop metrics, following the official definition~\cite{Bench2Drive}. 

Success Rate (SR): This metric measures the proportion of successfully completed routes within the allotted time and without traffic violations. A route is deemed successful if the ego vehicle reaches its destination without any rule infractions. The success rate is calculated as the ratio of successful routes to the total number of routes, as shown below: 

\begin{equation}
    \text{SR} = \frac{n_{\text{success}}}{n_{\text{total}}}.  
\end{equation}

Driving Score (DS): This metric is based on the official CARLA~\cite{carla} evaluation protocol. It considers both route completion and penalty for infractions. Specifically, it averages the route completion percentages and penalizes infractions based on their severity, as depicted below. The driving score is normalized by the total number of routes from the same type or group as well.

\begin{equation}
    \text{DS} = \frac{1}{n_{\text{total}}} \sum\limits_{i=1}^{n_{\text{total}}} \text{Route-Completion}_i * \prod\limits_{j=1}^{n_{i, \text{penalty}}} p_{i, j},
\end{equation}

where $n_{\text{success}}$ and ${n_{\text{total}}}$ denote the number of successful routes and total samples respectively; $\text{Route-Completion}_i$ represents the percentage of route distance completed for the $i$-th route; $p_{i,j}$ means the $j$-th infraction penalty on the $i$-th route.

\section{Notations}
As shown in Table~\ref{tab:notation}, we provide a lookup table of the notations used in the paper.


\begin{table} [ht!] 

\centering
{\begin{tabular}{c|l}
\toprule[1.5pt]
Notation & Description \\
[1.5pt]\hline\noalign{\vskip 2pt}

$I$ & multi-view images \\
$N_{\rm img}$ & the number of camera views \\
$F_{\rm img}$ & multi-view image features \\
$F_{\rm BEV}$ & bird’s-eye-view (BEV) features \\
$F_{\rm agent}$ & surrounding agent features \\
$Q_{\rm traj}$ & multi-mode trajectory queries \\
$M$ & the number of planning modes \\
$C$ & the feature channels \\
$P_{\rm t}$ & guiding points at time $t$ \\
$F_{\rm DisRel}$ & the relative distance features \\
$F_{\rm AgtRel}$ & the distance-aware agent features \\
$\Delta P_{\rm ft}$ & planning trajectory offsets \\
$P_{\rm ft}$ & planning trajectory points \\
$P_{\rm f}$ & planning trajectories \\
$S_{\rm f}$ & planning classification score \\

\bottomrule[1.5pt]
\end{tabular}}
\caption{Notations used in the paper.}

\label{tab:notation}
\end{table}

\section{Qualitative results}
\begin{figure*}[ht!]
\centering

\includegraphics[width=0.6\textwidth]{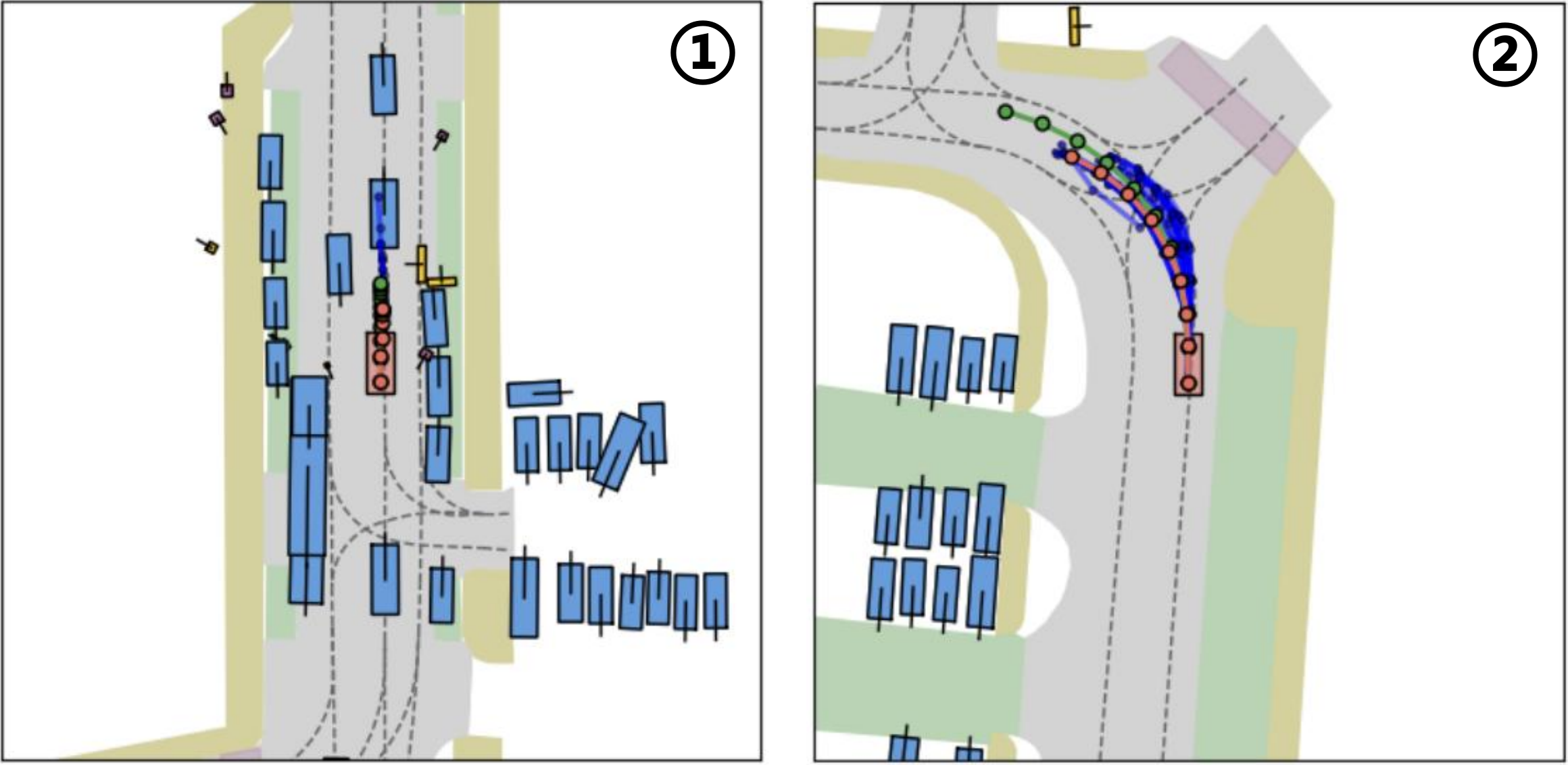}

\caption{
Qualitative results \textit{on the NAVSIM~\cite{navsim} dataset}. The ground-truth trajectory is shown in green, while the best-planned trajectory, selected based on the classification score, is highlighted in orange.
}
\label{fig:supp_navsim}
\end{figure*}

\begin{figure*}[ht!]
\centering
\includegraphics[width=1\linewidth]{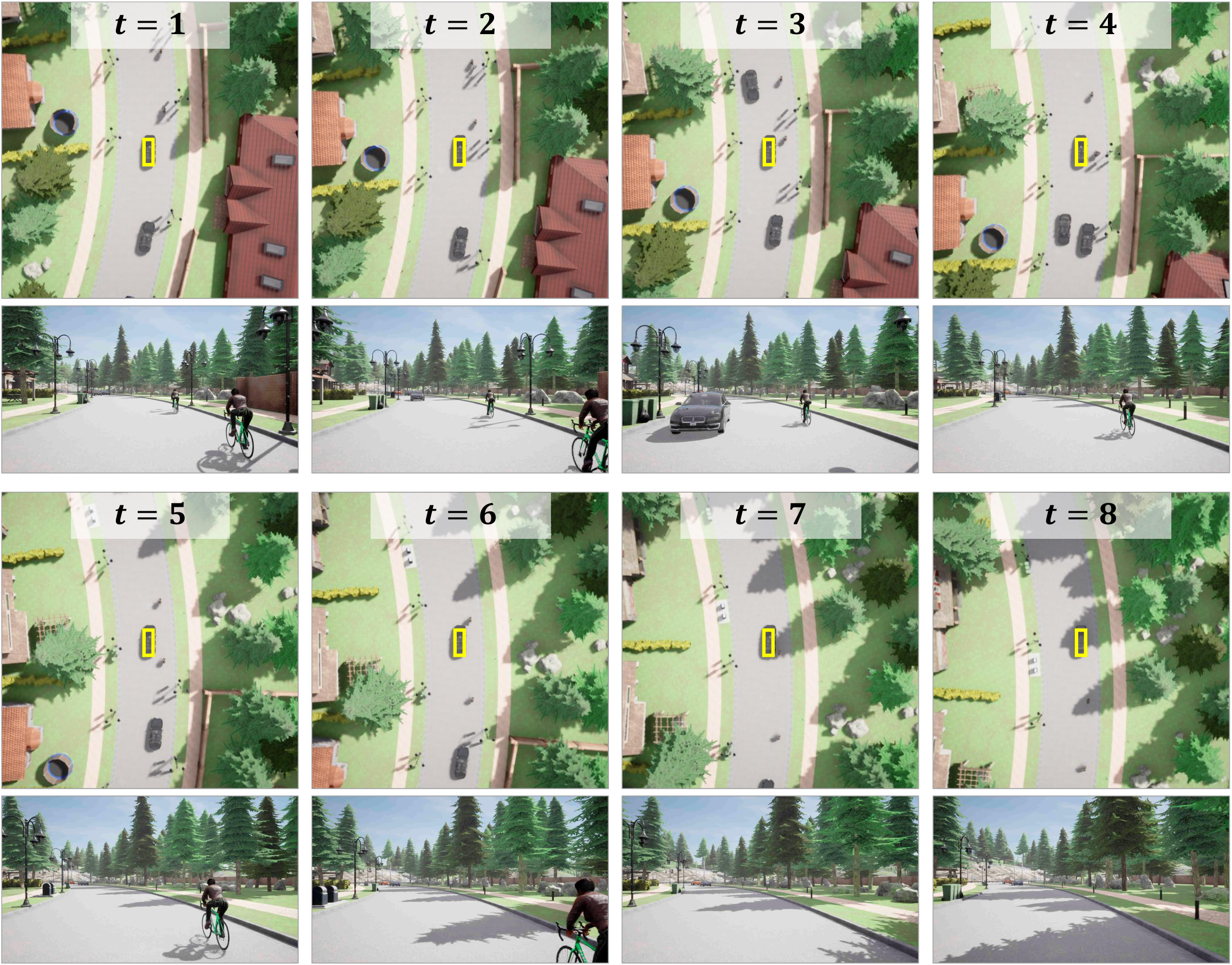}

\caption{
Qualitative results \textit{on the Bench2Drive~\cite{Bench2Drive} dataset}. The upper image shows the BEV view, while the lower image presents the front view. In the BEV view, the ego vehicle is marked with a yellow box. In this case, our model successfully steers to the left to avoid cyclists on the right side of the road.
}
\label{fig:vis_main_bench}
\end{figure*}

We present additional qualitative results of our \netName{} in Figure~\ref{fig:supp_navsim} and Figure~\ref{fig:vis_main_bench}, based on the NAVSIM~\cite{navsim} and Bench2Drive~\cite{Bench2Drive} datasets, respectively.
As shown in Figure~\ref{fig:supp_navsim}, our model generates trajectories that align closely with the ground truth.
In Figure~\ref{fig:vis_main_bench}, the model successfully steers to the left to avoid cyclists positioned on the right side of the road, demonstrating its ability to handle complex dynamic scenes.

\begin{figure*}[t!]
\centering

\includegraphics[width=1\textwidth]{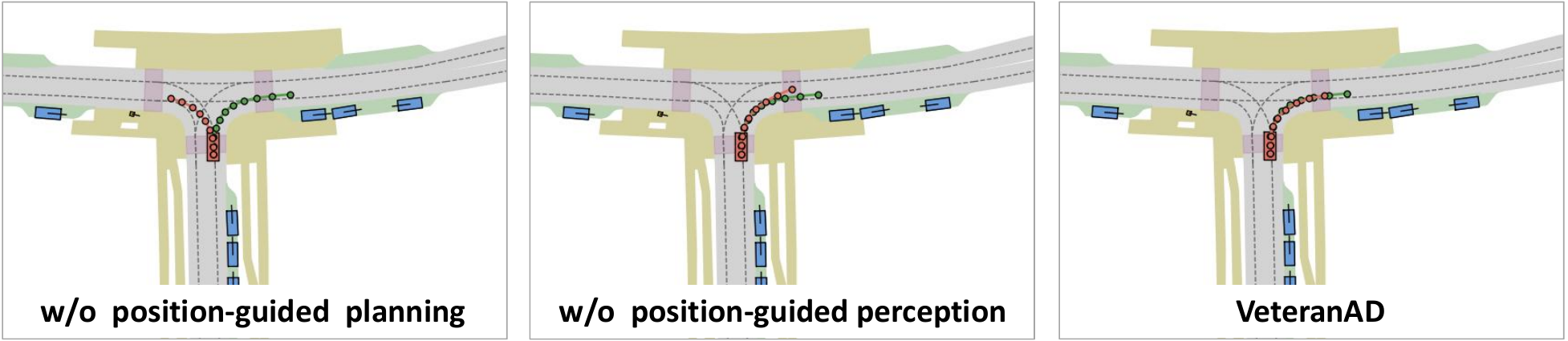}
\caption{
Visual results of the ablation study on model design. The left image shows the planning results without position guidance in the Localized Autoregressive Trajectory Planning module; the middle image shows results without position guidance in the Planning-Aware Holistic Perception module; and the right image presents the full model output of our \netName{}. The ground-truth trajectory is shown in green, while the predicted trajectory is highlighted in orange.
}
\label{fig:rebuttal}
\end{figure*}

As shown in Figure~\ref{fig:rebuttal}, we conduct an ablation study with visual results to demonstrate the effectiveness of our \netName{}. In the left image, without position-guided planning, the model must generate trajectories entirely from scratch instead of adjusting anchored trajectories, leading to incorrect planning outcomes. In the middle image, without position-guided perception, the model fails to perform targeted perception and thus struggles to capture detailed map features, resulting in inaccurate planning. In contrast, the right image shows that our \netName{}, equipped with the Localized Autoregressive Trajectory Planning module for localized planning and the Planning-Aware Holistic Perception module for targeted perception, produces accurate and reliable planning results.
This highlights the effectiveness of our design.

\section{Failure cases}

\begin{figure*}[ht!]
\centering

\includegraphics[width=1\textwidth]{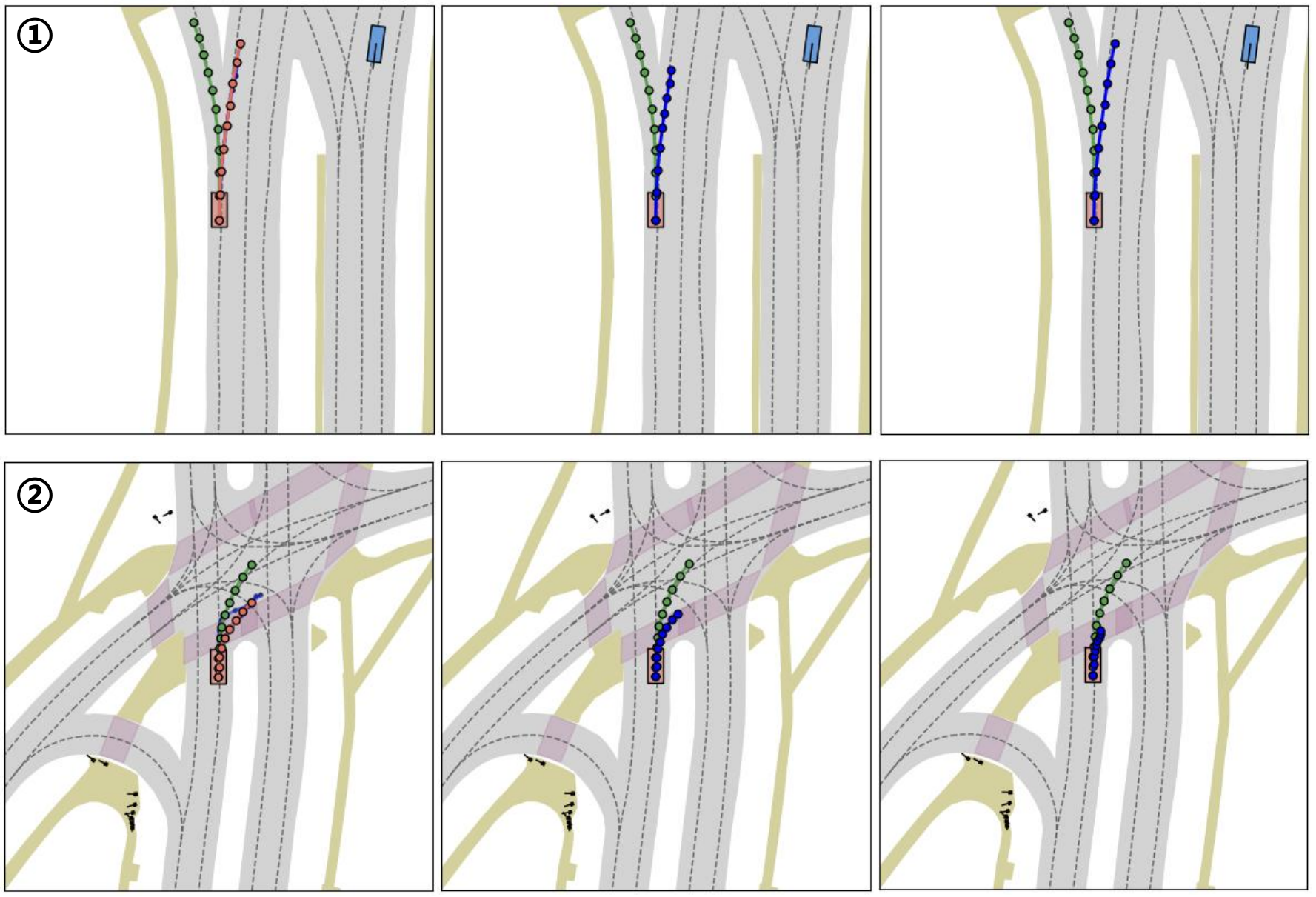}

\caption{
Failure cases \textit{on the NAVSIM~\cite{navsim} dataset}. The ground-truth trajectory is depicted in green, while the best planning trajectory, selected based on the classification score, is shown in orange. Additional planning trajectory modes are illustrated in blue.
}
\label{fig:supp_fial_nav}
\end{figure*}

\begin{figure*}[ht!]
\centering

\includegraphics[width=1\textwidth]{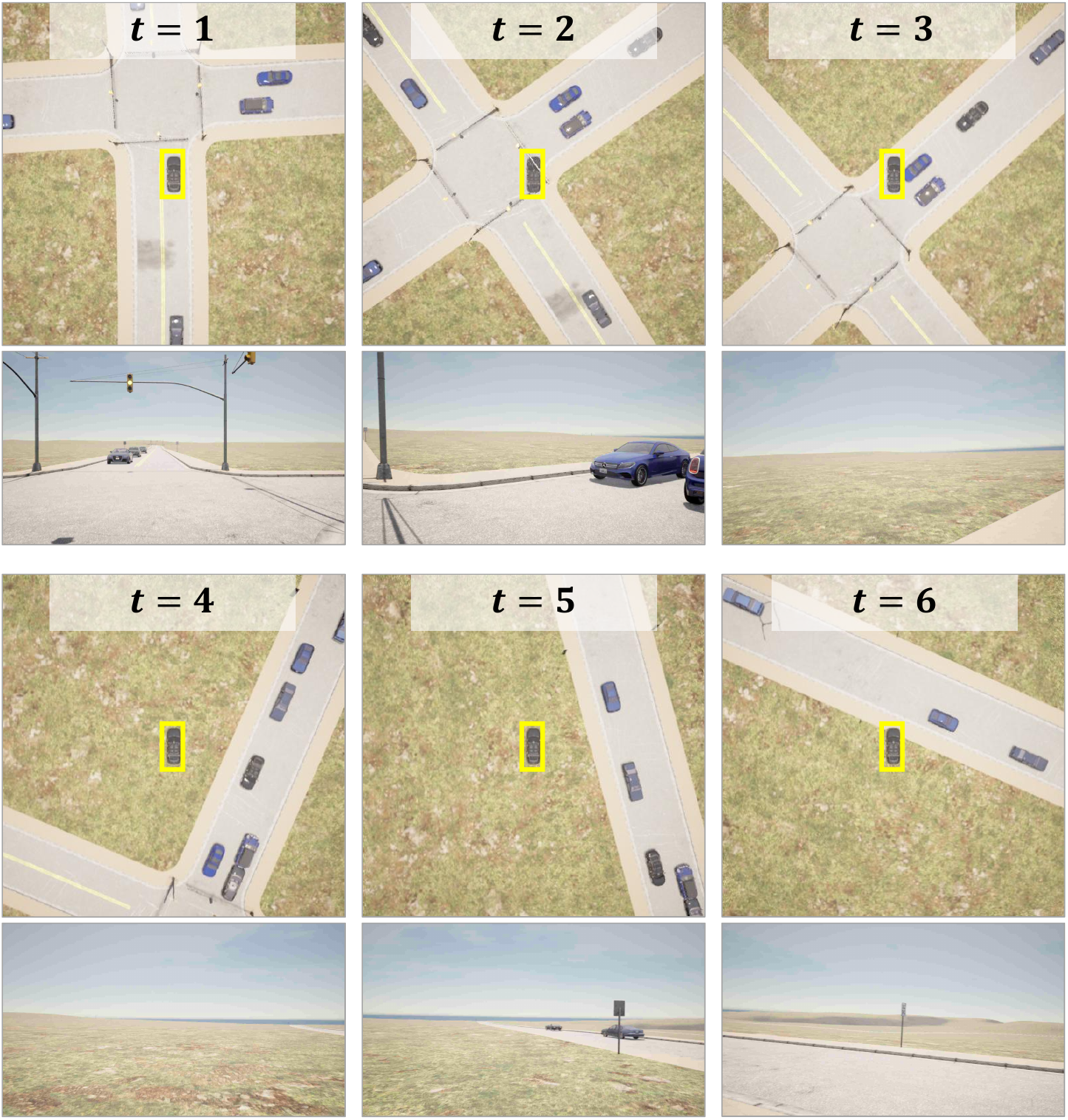}

\caption{
Failure cases \textit{on the Bench2Drive~\cite{Bench2Drive} dataset} in the closed-loop simulation. The upper image shows the BEV view, while the lower image presents the front view. In the BEV view, the ego vehicle is marked with a yellow box.
}
\label{fig:supp_fial_ben}
\end{figure*}

Although our \netName{} demonstrates strong performance on end-to-end planning benchmarks, it still exhibits certain failure cases. Representative examples from the NAVSIM~\cite{navsim} and Bench2Drive~\cite{Bench2Drive} datasets are shown in Figure~\ref{fig:supp_fial_nav} and Figure~\ref{fig:supp_fial_ben}, respectively.

In Figure~\ref{fig:supp_fial_nav}, the first case shows the model selecting an incorrect road for planning, which could potentially be addressed by incorporating explicit navigation instructions. In the second case, although the model successfully captures the right-turn intention, the predicted path lacks precision. This issue may be alleviated by improving the model's sensitivity to lane directions.

In Figure~\ref{fig:supp_fial_ben}, the model fails in a closed-loop simulation scenario, veering off the road during a right turn. This highlights a limitation of purely learning-based planning approaches under closed-loop settings, where the absence of explicit traffic rules may lead to infeasible behaviors. Incorporating rule-based constraints or traffic regulation priors could help mitigate such failures.


\bibliography{aaai2026}

\end{document}